\documentclass[11pt,a4paper]{article}
\usepackage[hyperref]{acl2021}
\usepackage{times}
\usepackage{latexsym}

\usepackage{comment}
\usepackage{graphicx}
\usepackage{subcaption}
\usepackage{booktabs}
\usepackage{multirow}
\usepackage[hang,flushmargin]{footmisc}

\usepackage{microtype}

\aclfinalcopy 
\def\aclpaperid{1860} 

\title{Improving Compositional Generalization in Classification Tasks \\ via Structure Annotations}
  
\author{Juyong Kim ~~ Pradeep Ravikumar \\
  Carnegie Mellon University \\
  \small\texttt{\{juyongk,pradeepr\}@cs.cmu.edu} \\
\And
  Joshua Ainslie ~~ Santiago Onta\~{n}\'{o}n \\
  Google Research \\
  \small\texttt{\{jainslie,santiontanon\}@google.com}
\\}

\date{}

\begin{document}
\maketitle
\begin{abstract}
Compositional generalization is the ability to generalize systematically to a new data distribution by combining known components. Although humans seem to have a great ability to generalize compositionally, state-of-the-art neural models struggle to do so. In this work, we study compositional generalization in classification tasks and present two main contributions. First, we study ways to convert a natural language sequence-to-sequence dataset to a classification dataset that also requires compositional generalization. Second, we show that providing structural hints (specifically, providing parse trees and entity links as attention masks for a Transformer model) helps compositional generalization. 
\end{abstract}

\section{Introduction}

{\em Compositional generalization} is the ability of a system to systematically generalize to a new data distribution by combining known components or primitives. For example, assume a system has learned the meaning of ``jump'' and that ``jump twice'' means that the action ``jump'' has to be repeated two times. Upon learning the meaning of the action ``jax'', it should be able to infer what ``jax twice'' means. 
Although modern neural architectures are pushing the state of the art in many complex natural language tasks, these models still struggle with compositional generalization~\cite{hupkes2020compositionality}.



In order to advance research in this important direction, in this paper we present two main contributions \footnote{\fontsize{8.5pt}{10.2pt} \url{http://goo.gle/compositional-classification}}.
First, we present a binary classification dataset which is hard in a compositional way. This allows for studying the compositional generalization ability of a larger range of models than sequence generation tasks, since the task only requires an encoder, and not a decoder. Specifically, we present a methodology to convert an existing semantic parsing dataset, CFQ~\cite{keysers2019measuring}, into a binary classification dataset that is also compositionally hard.

Our second and main contribution is showing that a transformer-based model can better generalize compositionally if we provide hints on the structure of the input. Specifically, we do so by modifying the attention mask used by the model. This is an interesting result, as (except for two additions, which we elaborate on in Section \ref{sec:annotations}) attention masks do not ``add'' any attention capabilities to the model. Instead, it seems that it is the removal of certain attention pairs that makes the difference. This suggests that vanilla Transformer is having a hard time suppressing non-compositional attention.


\section{Background}

This section overviews existing work on compositional generalization and then some background on the Transformer models used in this paper.
Please see Section \ref{sec:related_works} in the appendix for detailed review.


{\bf Compositional Generalization.}
Compositional generalization can manifest in different ways~\cite{hupkes2020compositionality} such as {\em systematicity} (recombination of known parts and rules) or {\em productivity} (extrapolation to longer sequences than those seen during training), among others.
%
%
Early work focused on showing how different deep learning models do not generalize compositionally~\cite{livska2018memorize}, and datasets such as SCAN~\cite{lake2018generalization} 
or CFQ~\cite{keysers2019measuring} were proposed to show these effects. 


Work toward improving compositional generalization has proposed ideas such as Syntactic attention~\cite{russin2019compositional}, 
increased pre-training~\cite{furrer2020compositional},
data augmentation~\cite{andreas2019good}, 
or general purpose sequential models such as {\em Neural Turing Machines} or {\em Differential Neural Computers}~\cite{graves2016hybrid}.





{\bf ETC.} 
For our experimental evaluation we use the ETC~\cite{ainslie2020etc} Transformer model. ETC extends the standard Transformer model in 3 key ways: (1) it uses a global-local attention mechanism to scale to long inputs, (2) it uses relative attention~\cite{shaw2018self} and flexible masking and (3) it uses a new pre-training loss based on Contrastive Predictive Coding (CPC)~\cite{oord2018representation}. The last two extensions allow it to handle structured inputs containing, for example, hierarchical structure. In this work, we rely on (2) to annotate the structure of the input.


\section{The CFQ Classification Dataset}

\begin{figure*}[t]
    \centering
    \begin{minipage}{0.33\textwidth}
        \centering
        \includegraphics[width=\textwidth]{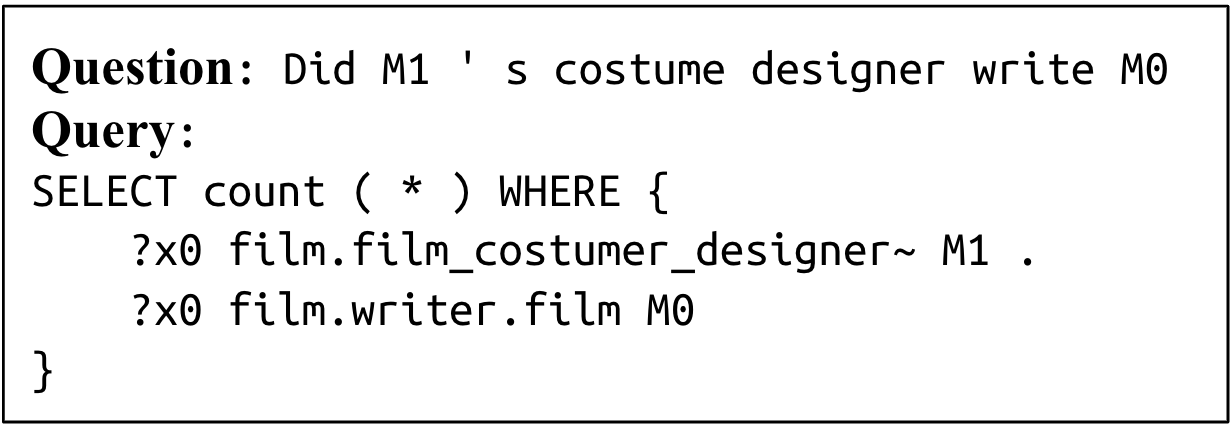}
        \subcaption{A CFQ example}
        \label{fig:struct_anno_ex}
        \vspace{0.75em}
        \includegraphics[width=\textwidth]{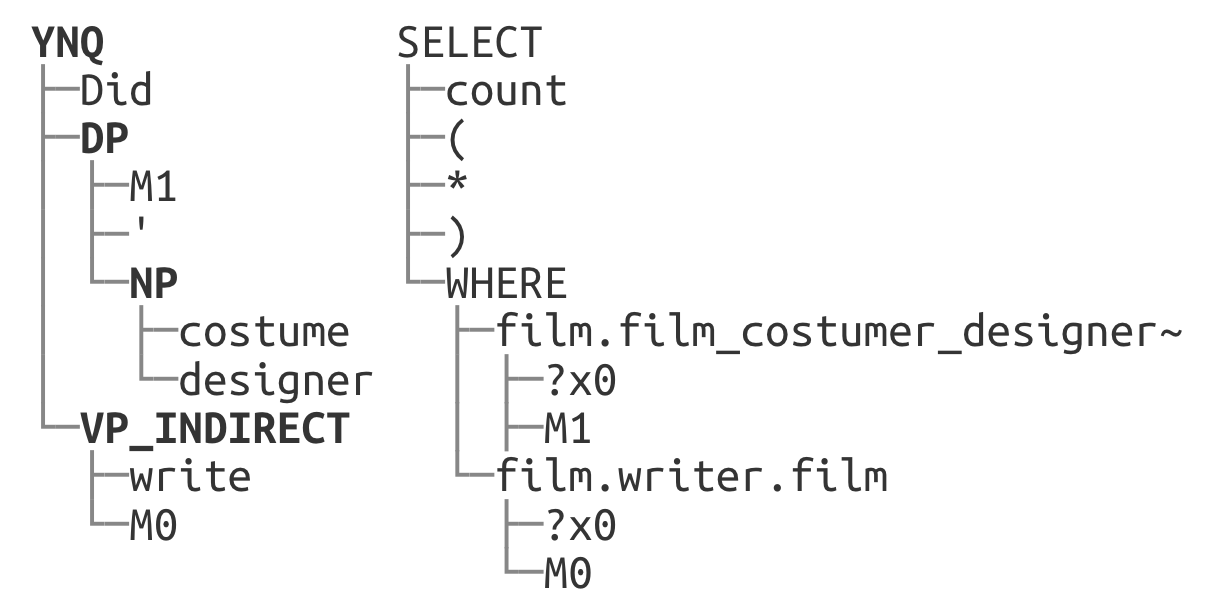}
        \subcaption{Parse trees of the CFQ example}
        \label{fig:struct_anno_tree}
    \end{minipage}
    \hspace{-0.5em}
    \begin{minipage}{0.33\textwidth}
        \centering
        \includegraphics[width=\textwidth]{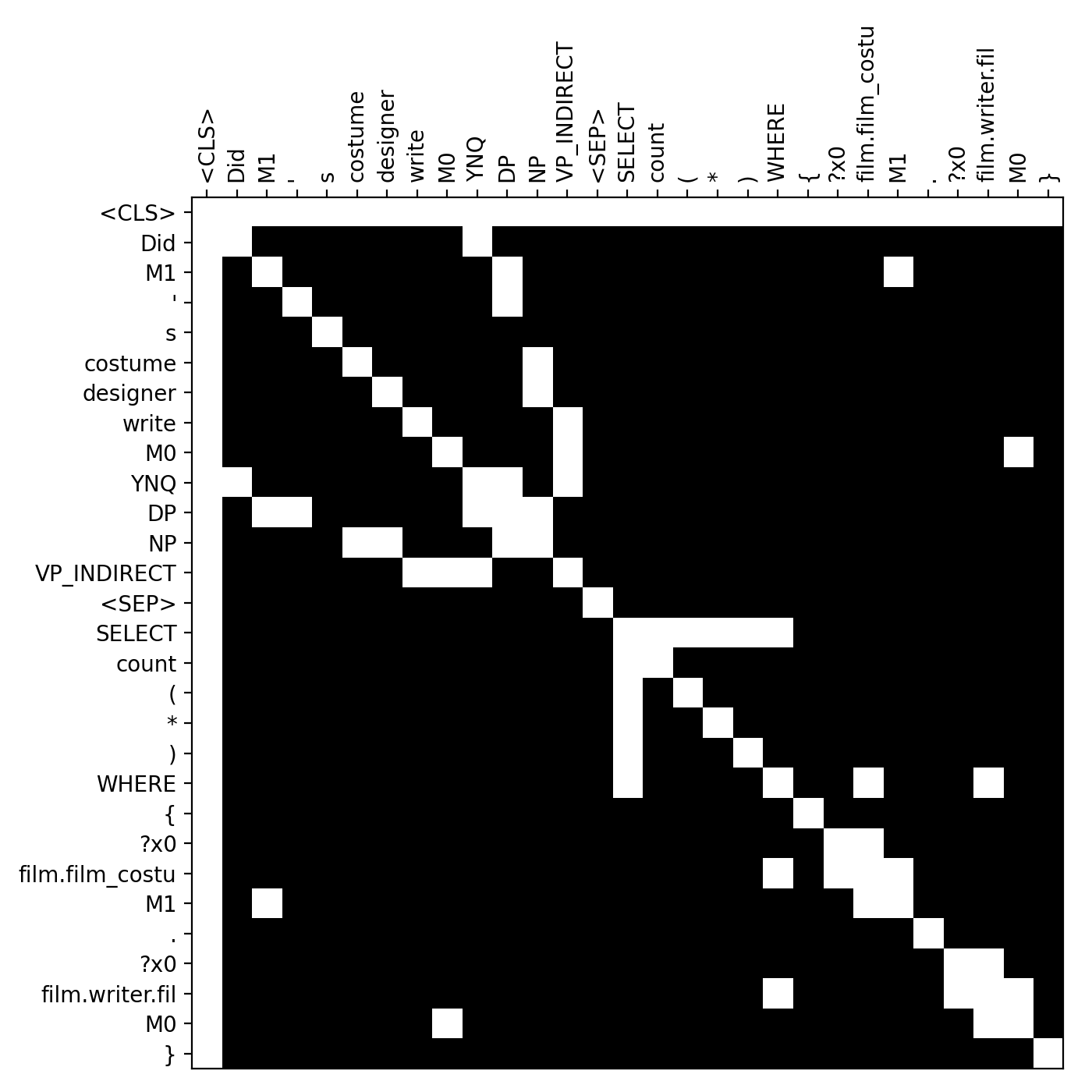}
        \subcaption{Hard mask}
        \label{fig:struct_anno_hard}
    \end{minipage}
    \hspace{-0.5em}
    \begin{minipage}{0.33\textwidth}
        \centering
        \includegraphics[width=\textwidth]{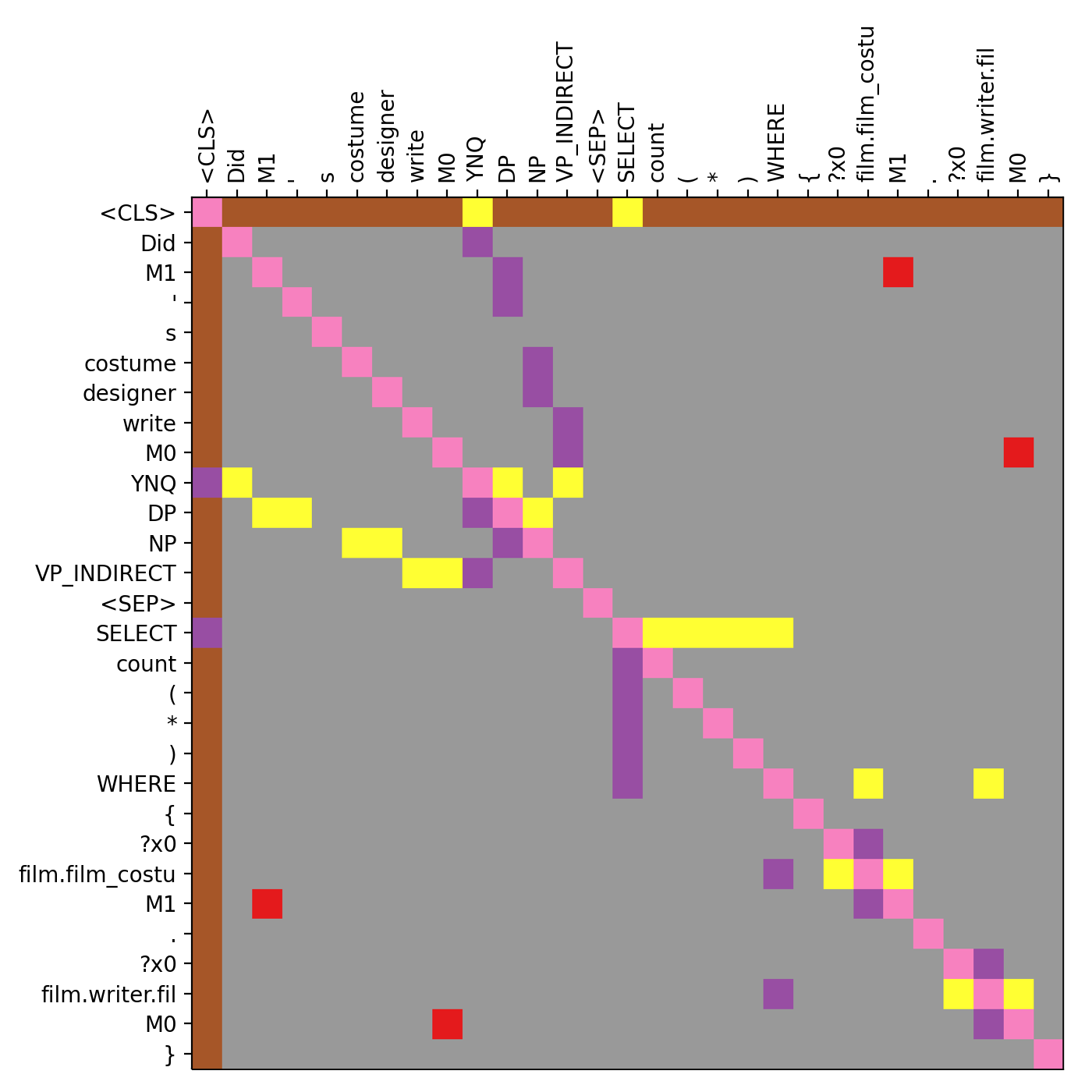}
        \subcaption{Soft mask}
        \label{fig:struct_anno_soft}
    \end{minipage}
    \caption{Structure annotations for a CFQ example. We extract the hierarchical structure of the question and query of CFQ examples and use them to mask attention (hard mask) and/or provide relative attention labels (soft). Different colors indicate different relative attention labels.}
    \label{fig:struct_anno}
\end{figure*}

The Compositional Freebase Questions (CFQ) dataset \citep{keysers2019measuring} is an NLU dataset to measure the compositional capability of a learner.
It is designed around the task of \textit{translating} a natural language question into a SPARQL query.
The dataset has been automatically generated by a grammar 
and contains 239,357 sentence/query pairs.
An example is shown in Figure~\ref{fig:struct_anno_ex}.

As shown in the original work of \citeauthor{keysers2019measuring} \citeyearpar{keysers2019measuring} in order to properly measure the compositional generalization ability of a model, the train and test sets should be split with similar distributions of tokens ({\em atoms}), but different distributions of their compositions (the {\em compounds}).
In the CFQ dataset, to ensure this, two divergences, namely {\em atom divergence} and {\em compound divergence}, between the train and dev/test set are measured while constructing the splits.
As a result, carefully selected splits called {\em maximum compound divergence (MCD)} splits are hard for standard neural networks (they perform well in the train set, but poorly in the test set), while the random splits are easier.

We convert the CFQ dataset into a dataset with a binary classification task. In this new dataset, the input is a question and a SPARQL query, and the task is to determine whether these two sequences have the same meaning or not.
Two considerations must be made to ensure the resulting dataset requires compositional generalization:

\begin{figure*}[t]
    \centering
    \includegraphics[width=\textwidth]{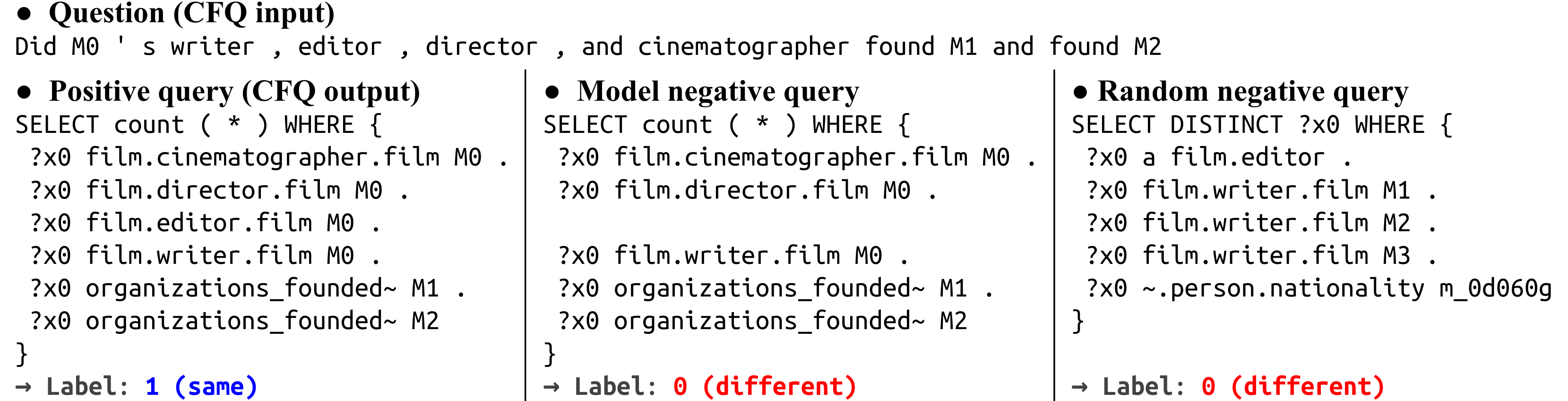}
    \caption{Examples of the CFQ classification dataset. Each query pairs with the question to form an instance. Note the model negative resembles the positive, while the random negative query differs considerably. 
    }
    \label{fig:cfq_cls_example}
\end{figure*}

\begin{figure}[t]
    \centering
    \includegraphics[width=\columnwidth]{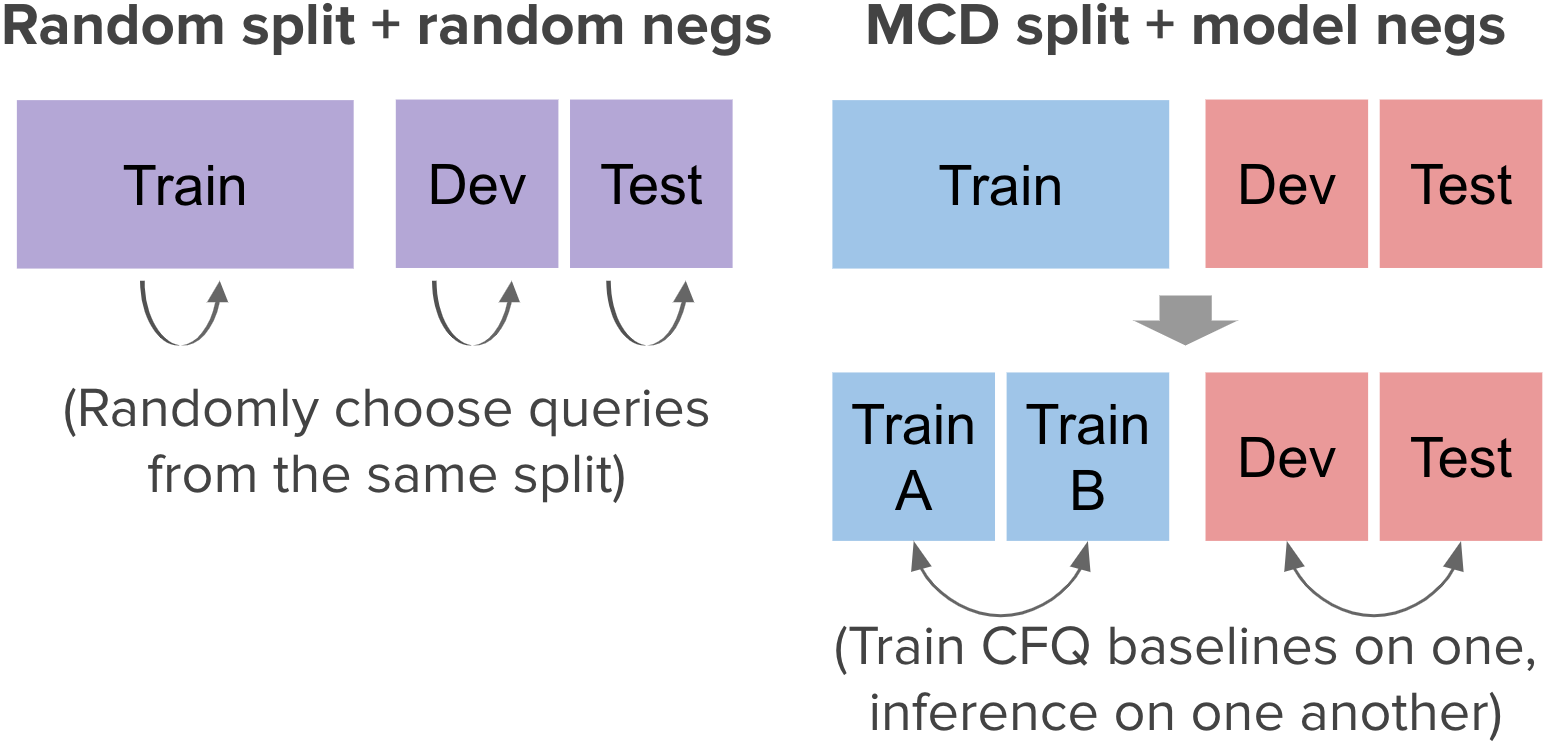}
    \caption{Negative example strategies. Different colors indicate different compound distributions.}
    \label{fig:neg_strategy}
\end{figure}

\paragraph{Negative Example Strategies:}
Positive instances of the binary classification task can be obtained directly from the original dataset, but 
to obtain negatives, we use either of two strategies:

\begin{itemize}
    \item Random negatives: We pair each question with a randomly chosen query. 
    \item Model negatives: Using baseline models (LSTM~\cite{hochreiter1997long},  Transformer~\cite{vaswani2017attention}, and Universal Transformer~\cite{dehghani2018universal}) trained on the original CFQ dataset, we get top-$k$ query predictions for each question. After filtering syntactically invalid queries and duplicates, we can get hard examples for classification from their incorrect predictions.
\end{itemize}

Model negatives are important, as otherwise, the task becomes too easy and would likely not require compositional generalization. See Figure~\ref{fig:cfq_cls_example} for examples of random/model negative instances.

\paragraph{Compound Distribution of Negative Examples:}
To 
prevent data leakage (e.g., compounds from the test set of the original CFQ dataset leaking into the training set of the classification CFQ dataset), we carefully choose the sampling set for random negatives and the train and inference set for model negatives.
We generate two splits of the original CFQ dataset. Each split contains three sets with 50\% data on train, 25\% on dev and 25\% on test. The first is a {\em random split} of the data, and the second ({\em MCD split}), maximizes the compound divergence between train and dev/test using the same method as in the original CFQ work. Then, we process the examples in each of these sets generating positive and negative examples. For random negatives, we sample negative queries for each questions from the set which the original example belongs to (train/dev/test).
For model negatives, to generate negatives for the training set, we divide it into two halves, train models in one, and generate negatives with the other half. For dev/test, we train on dev and generate negatives on test, and vice versa.
Figure~\ref{fig:neg_strategy} illustrates this procedure, designed to ensure there is no leakage of compounds between train and dev/test.

For both strategies, we make 1 positive and 3 negatives per original CFQ example.
Also, we set aside 5\% of the train set as a hold-out set to check i.i.d.\ generalization.

\section{Compositional Generalization via Structure Annotation}\label{sec:annotations}

Our hypothesis is that part of the difficulty in compositional generalization is to parse the structure of the input. To test this, we evaluate the performance of models when we provide annotations for two structural elements of the inputs: parse trees of both the natural language sentences and SPARQL queries, and {\em entity cross links} (linking entity mentions from the natural language side to the corresponding mentions in the SPARQL query). 

The parse trees of the questions are already given in the original CFQ dataset as constituency-based parse trees.
Since the trees include intermediate nodes indicating syntactic structures, we append tokens representing them at the end of each question. We created a simple parser to generate dependency-based parse trees for the SPARQL queries.
We join the roots of the two trees to make a single global tree with the \texttt{<CLS>} token as the root.

We represent the structure of the inputs by masking attention (``hard mask'') or with relative attention \citep{shaw2018self} labels (``soft mask'').

\begin{itemize}
    \item Hard mask: We customize the binary attention mask of the original Transformer to only allow attention between tokens connected by the edges of the parse tree.
    \item Soft mask: For every pair of input tokens, we assign relative attention labels based on which of the following edge relationships applies: parent-to-child, child-to-parent, itself, from-or-to-root, or entity-cross-link.
\end{itemize}

Additionally, we allow attention pairs in the masks connecting the entities appearing both in the question and the queries. We call these links \textit{entity cross links}, and they are found by simple string match (e.g. ``{\tt M0}''). Notice that while relative attention labels and the additional tokens to represent the constituency parse tree of the natural language add capabilities to the model, the ``hard mask'' structure annotations described above (which result in the larger performance gains) do not \textit{add} any attention capabilities to the model. Instead, they simply remove non-structure attention edges.
Figure~\ref{fig:struct_anno_tree} shows the parse trees, and Figure~\ref{fig:struct_anno_hard} and \ref{fig:struct_anno_soft} show the masks for an example.

\section{Results and Discussion}
\label{sec:results}


We used the ETC~\cite{ainslie2020etc} Transformer model implementation as it allows us to provide the hard and soft masks described above in an easy way.
In all experiments, we report AUC in the dev set as the evaluation metric (we did not evaluate on the test set). Please see Section~\ref{sec:exp_result_full} in the appendix for training details.

\subsection{The CFQ Classification Dataset}

\begin{table*}[t]
    \setlength{\tabcolsep}{10pt}
    \centering
    \begin{tabular}{c|ccc|ccc} \hline
         \multirow{2}{*}{\textbf{Model}}& \multicolumn{3}{c}{\textit{Random Split \& Random Neg}} & \multicolumn{3}{|c}{\textit{MCD Split \& Model Neg}} \\
          & \textbf{Train} & \textbf{Hold-out} & \textbf{Dev} & \textbf{Train} & \textbf{Hold-out} & \textbf{Dev} \\ \hline
         LSTM                   & 1.0000 & 0.9998 & 0.9998 & 1.0000 & 0.9972 & 0.8310 \\
         Transformer (2 layers) & 0.9998 & 0.9997 & 0.9998 & 0.9988 & 0.9931 & 0.8789 \\
         Transformer (6 layers) & 0.9999 & 1.0000 & 0.9999 & 0.9995 & 0.9931 & 0.8738 \\ \hline
    \end{tabular}
    \caption{AUC on the CFQ classification dataset generated with different methods}
    \label{tab:cfq_cls_result}
\end{table*}

\begin{table}[t]
    \resizebox{\columnwidth}{!}{%
    \centering
    \setlength{\tabcolsep}{2pt}
    \begin{tabular}{c|cc|ccc} \hline
        \multirow{2}{*}{\textbf{Model}} & \multirow{2}{*}{\shortstack{\textbf{Mask}\\\textbf{Type}}} & \multirow{2}{*}{\shortstack{\textbf{Cross}\\\textbf{link}}} & \multicolumn{3}{c}{\textit{MCD Split \& Model Neg}} \\
        & & & \textbf{Train} & \textbf{Hold-out} & \textbf{Dev} \\ \hline
        LSTM & \multicolumn{2}{c|}{-} & 1.0000 & 0.9972 & 0.8310 \\ \hline
        Transformer & \multicolumn{2}{c|}{-} & 0.9995 & 0.9931 & 0.8738 \\ \hline
        \multirow{5}{*}{\shortstack{Transformer\\ w/ structure\\annotations\\(ETC)}} & \multicolumn{1}{c|}{No} & - & 0.9994 & 0.9934 & 0.8868 \\ \cline{2-3}
         & \multirow{2}{*}{Hard} & \multicolumn{1}{|c|}{N} & 0.9999 & 0.9978 & 0.9061 \\ \cline{3-3}
         & \multicolumn{1}{c|}{} & \multirow{3}{*}{Y} & 1.0000 & 0.9992 & \underline{0.9656} \\ \cline{2-2}
         & \multicolumn{1}{c|}{Soft} & & 0.9995 & 0.9936 & 0.8819 \\ \cline{2-2}
         & \multicolumn{1}{c|}{Both} & & 1.0000 & 0.9991 & \textbf{0.9721} \\ \hline
    \end{tabular}
    }
    \caption{AUC on the CFQ classification dataset (\textit{MCD Split \& Model Neg}) with various structure annotations}
    \label{tab:strucu_anno_result}
\end{table}

We generate two classification datasets: ``random split \& random negatives'' and ``MCD split \& model negatives'', and evaluate LSTM and Transformer models.
For both datasets, we evaluate AUC on the hold-out set (taken out of the training set as described above) to test i.i.d.\ generalization, and on the dev set to test compositional generalization.

As shown in Table~\ref{tab:cfq_cls_result}, models easily generalize on the hold-out set (AUC $\geq$  0.99).
All baseline models also achieve almost 1.0 AUC in the dev set of the ``random split \& random negatives''.
However, in the ``MCD split \& model negatives'' models cannot generalize well on the dev set, showing compositional generalization is required. Note that random guessing achieves 0.5 AUC score.

\subsection{Structure Annotation}

Table \ref{tab:strucu_anno_result} compares different ablations of our structure annotation approach compared to the baseline models. The first (no masks and no cross links) just shows that adding tokens to the input to represent the constituency parsing and moving to ETC only provide small gains (from 0.8738 to 0.8868 AUC). Adding a hard mask already helps the model (0.9061 AUC), and adding cross links on top of that achieves very significant gains (0.9656 AUC). Finally, soft masks by themselves do not seem to help, but a combination of soft and hard masks achieves our best result of 0.9721 AUC.

The interesting result here is that adding the hard mask with entity cross links {\em only removes} potential attention pairs, so it does not increase model capacity in any way. In other words, the underlying transformer model is in principle able to generalize compositionally to some extent but seems to struggle in suppressing non-compositional attention.

\section{Conclusions}

The main contribution of this paper is to show that providing structure annotations in the form of attention masks significantly helps Transformer models generalize compositionally. This is interesting for two main reasons: first, it shows that neural network models do have the innate ability to generalize compositionally to some extent, but need some guidance to do so (e.g., by providing attention masks as in our work). This reinforces previous work showing that LSTMs also can, in principle, generalize compositionally, but they just do so with very low probability~\cite{livska2018memorize}. The second reason is that structure annotations, which we provided manually, could be generated by another model in future work. We also presented a procedure for generating classification datasets that require some degree of compositional generalization starting from sequence generation datasets.

\bibliographystyle{acl_natbib}

\begin{thebibliography}{15}
\expandafter\ifx\csname natexlab\endcsname\relax\def\natexlab#1{#1}\fi

\bibitem[{Abadi et~al.(2016)Abadi, Barham, Chen, Chen, Davis, Dean, Devin,
  Ghemawat, Irving, Isard et~al.}]{abadi2016tensorflow}
Mart{\'\i}n Abadi, Paul Barham, Jianmin Chen, Zhifeng Chen, Andy Davis, Jeffrey
  Dean, Matthieu Devin, Sanjay Ghemawat, Geoffrey Irving, Michael Isard, et~al.
  2016.
\newblock Tensorflow: A system for large-scale machine learning.
\newblock In \emph{12th $\{$USENIX$\}$ symposium on operating systems design
  and implementation ($\{$OSDI$\}$ 16)}, pages 265--283.

\bibitem[{Ainslie et~al.(2020)Ainslie, Onta\~{n}\'{o}n, Alberti, Cvicek,
  Fisher, Pham, Ravula, Sanghai, Wang, and Yang}]{ainslie2020etc}
Joshua Ainslie, Santiago Onta\~{n}\'{o}n, Chris Alberti, Vaclav Cvicek, Zachary
  Fisher, Philip Pham, Anirudh Ravula, Sumit Sanghai, Qifan Wang, and Li~Yang.
  2020.
\newblock {ETC}: Encoding long and structured inputs in transformers.
\newblock In \emph{Proceedings of the 2020 Conference on Empirical Methods in
  Natural Language Processing (EMNLP)}, pages 268--284.

\bibitem[{Andreas(2019)}]{andreas2019good}
Jacob Andreas. 2019.
\newblock Good-enough compositional data augmentation.
\newblock \emph{arXiv preprint arXiv:1904.09545}.

\bibitem[{Dehghani et~al.(2018)Dehghani, Gouws, Vinyals, Uszkoreit, and
  Kaiser}]{dehghani2018universal}
Mostafa Dehghani, Stephan Gouws, Oriol Vinyals, Jakob Uszkoreit, and {\L}ukasz
  Kaiser. 2018.
\newblock Universal transformers.
\newblock \emph{arXiv preprint arXiv:1807.03819}.

\bibitem[{Furrer et~al.(2020)Furrer, van Zee, Scales, and
  Sch{\"a}rli}]{furrer2020compositional}
Daniel Furrer, Marc van Zee, Nathan Scales, and Nathanael Sch{\"a}rli. 2020.
\newblock Compositional generalization in semantic parsing: Pre-training vs.
  specialized architectures.
\newblock \emph{arXiv preprint arXiv:2007.08970}.

\bibitem[{Graves et~al.(2016)Graves, Wayne, Reynolds, Harley, Danihelka,
  Grabska-Barwi{\'n}ska, Colmenarejo, Grefenstette, Ramalho, Agapiou
  et~al.}]{graves2016hybrid}
Alex Graves, Greg Wayne, Malcolm Reynolds, Tim Harley, Ivo Danihelka, Agnieszka
  Grabska-Barwi{\'n}ska, Sergio~G{\'o}mez Colmenarejo, Edward Grefenstette,
  Tiago Ramalho, John Agapiou, et~al. 2016.
\newblock Hybrid computing using a neural network with dynamic external memory.
\newblock \emph{Nature}, 538(7626):471--476.

\bibitem[{Hochreiter and Schmidhuber(1997)}]{hochreiter1997long}
Sepp Hochreiter and J{\"u}rgen Schmidhuber. 1997.
\newblock Long short-term memory.
\newblock \emph{Neural computation}, 9(8):1735--1780.

\bibitem[{Hupkes et~al.(2020)Hupkes, Dankers, Mul, and
  Bruni}]{hupkes2020compositionality}
Dieuwke Hupkes, Verna Dankers, Mathijs Mul, and Elia Bruni. 2020.
\newblock Compositionality decomposed: How do neural networks generalise?
\newblock \emph{Journal of Artificial Intelligence Research}, 67:757--795.

\bibitem[{Keysers et~al.(2019)Keysers, Sch{\"a}rli, Scales, Buisman, Furrer,
  Kashubin, Momchev, Sinopalnikov, Stafiniak, Tihon
  et~al.}]{keysers2019measuring}
Daniel Keysers, Nathanael Sch{\"a}rli, Nathan Scales, Hylke Buisman, Daniel
  Furrer, Sergii Kashubin, Nikola Momchev, Danila Sinopalnikov, Lukasz
  Stafiniak, Tibor Tihon, et~al. 2019.
\newblock Measuring compositional generalization: A comprehensive method on
  realistic data.
\newblock In \emph{International Conference on Learning Representations}.

\bibitem[{Lake and Baroni(2018)}]{lake2018generalization}
Brenden Lake and Marco Baroni. 2018.
\newblock Generalization without systematicity: On the compositional skills of
  sequence-to-sequence recurrent networks.
\newblock In \emph{International Conference on Machine Learning}, pages
  2873--2882. PMLR.

\bibitem[{Li{\v{s}}ka et~al.(2018)Li{\v{s}}ka, Kruszewski, and
  Baroni}]{livska2018memorize}
Adam Li{\v{s}}ka, Germ{\'a}n Kruszewski, and Marco Baroni. 2018.
\newblock Memorize or generalize? searching for a compositional rnn in a
  haystack.
\newblock \emph{arXiv preprint arXiv:1802.06467}.

\bibitem[{Oord et~al.(2018)Oord, Li, and Vinyals}]{oord2018representation}
Aaron van~den Oord, Yazhe Li, and Oriol Vinyals. 2018.
\newblock Representation learning with contrastive predictive coding.
\newblock \emph{arXiv preprint arXiv:1807.03748}.

\bibitem[{Russin et~al.(2019)Russin, Jo, O'Reilly, and
  Bengio}]{russin2019compositional}
Jake Russin, Jason Jo, Randall~C O'Reilly, and Yoshua Bengio. 2019.
\newblock Compositional generalization in a deep seq2seq model by separating
  syntax and semantics.
\newblock \emph{arXiv preprint arXiv:1904.09708}.

\bibitem[{Shaw et~al.(2018)Shaw, Uszkoreit, and Vaswani}]{shaw2018self}
Peter Shaw, Jakob Uszkoreit, and Ashish Vaswani. 2018.
\newblock Self-attention with relative position representations.
\newblock In \emph{Proceedings of the 2018 Conference of the North American
  Chapter of the Association for Computational Linguistics: Human Language
  Technologies, Volume 2 (Short Papers)}, pages 464--468.

\bibitem[{Vaswani et~al.(2017)Vaswani, Shazeer, Parmar, Uszkoreit, Jones,
  Gomez, Kaiser, and Polosukhin}]{vaswani2017attention}
Ashish Vaswani, Noam Shazeer, Niki Parmar, Jakob Uszkoreit, Llion Jones,
  Aidan~N Gomez, Lukasz Kaiser, and Illia Polosukhin. 2017.
\newblock Attention is all you need.
\newblock \emph{arXiv preprint arXiv:1706.03762}.

\end{thebibliography}

\clearpage

\appendix

\section{Full Experimental Results}
\label{sec:exp_result_full}

In this section, we report the full results on the CFQ classification dataset and the structure annotation experiments.
In all configurations, multiple evaluation metrics (accuracy, F1 score, and AUC) are computed by averaging the results of two randomly initialized experiments.
We test each network using only the val set, not the test set, since the main purpose of the experiment is to compare the compositional generalization ability, not to select best hyper-parameter.
Accuracy and F1 score are computed with the threshold 0.5 of the softmax output of label 1.

All the experiments of the CFQ classification datasets were run using the TensorFlow~\citep{abadi2016tensorflow} framework.
As we explain in the Section~\ref{sec:results}, we use the ETC Transformer~\cite{ainslie2020etc} code for relative position embeddings.
For the Transformer implementation, we use the code provided in a Tensorflow tutorial.
The training is run on the \texttt{n1-highmem-8} instance (52GB RAM, 8 virtual cpus) of Google Cloud Platform, extended with NVIDIA Tesla V100 GPUs.

Hyper-parameters used in the training of neural networks are listed in Table~\ref{tab:hyper_params}.
One thing that we want to clarify is that training steps are required number of steps to converge and the training did not last longer than needed.
Nevertheless, the experiments with structure annotations required more training steps than LSTM/Transformer, especially when the network is using hard mask.
We conjecture that training with the hard mask of parse trees is slow since only a small part of the attention is not masked and hence propagating the gradient via supervision at the \texttt{<CLS>} position is slow.

\begin{table*}[t]
    \centering
    \begin{tabular}{lccc} \hline
                               & \textbf{LSTM}      & \textbf{Transformer} & \textbf{ETC} \\ \hline
        Hidden layers          & 2                  & \{2,6\}           & 6                    \\
        Last dense layers      & 2                  & 1                 & 1                    \\
        Hidden Size            & 512                & 128               & 128                  \\
        Filter size            & -                  & 2048              & 512                  \\
        Number of heads        & -                  & 16                & 16                   \\
        Dropout                & 0.4                & 0.1               & 0.1                  \\
        Batch size             & 1024               & 512               & 112                  \\
        Training steps         &                    &                   &                      \\
        ~~~\textit{Random \& Random} & 20k                & 10k               & -                    \\
        ~~~\textit{MCD \& Random}    & 20k                & 10k               & -                    \\
        ~~~\textit{MCD \& Model}     & 30k                & 20k               & 200k                 \\
        Optimizer              & Adam (0.85, 0.997) & Adam (0.9, 0.997) & Adam (0.9, 0.997)    \\
        Learning rate schedule & Constant           & Inverse sqrt      & Inverse sqrt         \\
        Base learning rate     & 0.001              & 0.001             & 0.001                \\
        Warmup steps           & -                  & 1000              & 1000                 \\
        Weight decay           & 0.0                & 0.0               & 0.0                 \\ \hline
    \end{tabular}
    \caption{Hyper-parameters used in training deep neural networks on the CFQ classification datasets}
    \label{tab:hyper_params}
\end{table*}

\subsection{The CFQ classification Dataset}

Table~\ref{tab:cfq_cls_result_full} shows the classification results of various methods of generating classification datasets, including one additional configuration (\textit{MCD Split \& Random Negatives}).
The dataset generated by this new configuration has the train and the dev/test set that have different compound distributions, because it is based on the MCD split.
However, because of the method used in generating negative instances (random negatives), the binary classification of correspondence can be easily generalizable to the dev set.

\begin{table*}[t]
    \setlength{\tabcolsep}{4pt}
    \centering
    \begin{tabular}{c|ccc|ccc|ccc}
        \multicolumn{10}{c}{Dataset 1: \textit{Random Split \& Random Negatives}} \\ \hline
        \multirow{2}{*}{\textbf{Model}} & \multicolumn{3}{c|}{\textbf{Train}} & \multicolumn{3}{c|}{\textbf{Train (hold-out)}} & \multicolumn{3}{c}{\textbf{Dev}} \\
        & Acc & F1 & AUC & Acc & F1 & AUC & Acc & F1 & AUC \\ \hline
        LSTM                   & 0.9999 & 0.9998 & 1.0000 & 0.9984 & 0.9967 & 0.9998 & 0.9982 & 0.9964 & 0.9998 \\
        Transformer (2 layers) & 0.9988 & 0.9976 & 0.9998 & 0.9982 & 0.9964 & 0.9997 & 0.9988 & 0.9975 & 0.9998 \\
        Transformer (6 layers) & 0.9992 & 0.9988 & 0.9999 & 0.9989 & 0.9978 & 0.9999 & 0.9990 & 0.9979 & 0.9999 \\ \hline
        \multicolumn{10}{c}{Dataset 2: \textit{MCD Split \& Random Negatives}} \rule{0pt}{2.6ex} \\ \hline
        \multirow{2}{*}{\textbf{Model}} & \multicolumn{3}{c|}{\textbf{Train}} & \multicolumn{3}{c|}{\textbf{Train (hold-out)}} & \multicolumn{3}{c}{\textbf{Dev}} \\
        & Acc & F1 & AUC & Acc & F1 & AUC & Acc & F1 & AUC \\ \hline
        LSTM                   & 0.9999 & 0.9998 & 1.0000 & 0.9982 & 0.9965 & 0.9999 & 0.9546 & 0.9025 & 0.9923 \\
        Transformer (2 layers) & 0.9982 & 0.9965 & 1.0000 & 0.9974 & 0.9948 & 0.9999 & 0.9942 & 0.9883 & 0.9996 \\
        Transformer (6 layers) & 0.9986 & 0.9972 & 0.9999 & 0.9979 & 0.9958 & 0.9997 & 0.9889 & 0.9775 & 0.9991 \\ \hline
        \multicolumn{10}{c}{Dataset 3: \textit{MCD Split \& Model Negatives}} \rule{0pt}{2.6ex} \\ \hline
        \multirow{2}{*}{\textbf{Model}} & \multicolumn{3}{c|}{\textbf{Train}} & \multicolumn{3}{c|}{\textbf{Train (hold-out)}} & \multicolumn{3}{c}{\textbf{Dev}} \\
        & Acc & F1 & AUC & Acc & F1 & AUC & Acc & F1 & AUC \\ \hline
        LSTM                   & 0.9990 & 0.9979 & 1.0000 & 0.9796 & 0.9604 & 0.9972 & 0.8226 & 0.5199 & 0.8310 \\
        Transformer (2 layers) & 0.9817 & 0.9639 & 0.9988 & 0.9592 & 0.9202 & 0.9931 & 0.8359 & 0.5835 & 0.8789 \\
        Transformer (6 layers) & 0.9886 & 0.9776 & 0.9995 & 0.9582 & 0.9189 & 0.9931 & 0.8414 & 0.6191 & 0.8738 \\ \hline
    \end{tabular}
    \caption{Results of the CFQ classification dataset generated with different CFQ splits and negative example strategies}
    \label{tab:cfq_cls_result_full}
\end{table*}

\subsection{Structure Annotation}

\begin{figure}[t]
    \centering
    \includegraphics[width=0.8\columnwidth]{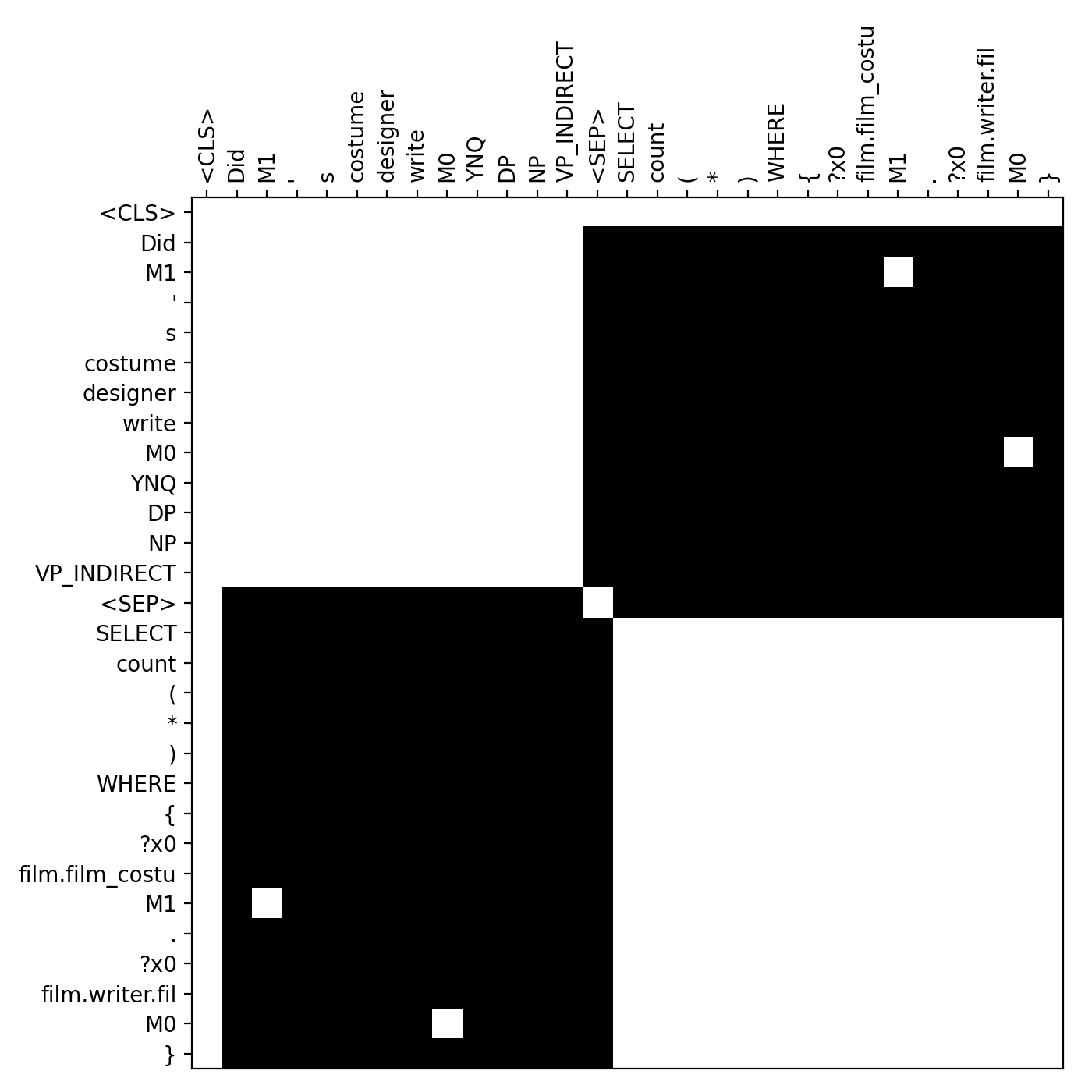}
    \caption{Block attention mask for the CFQ classification example of Figure~\ref{fig:struct_anno}. The dots at top-right and bottom-left are from entity cross links.}
    \label{fig:block_attn}
\end{figure}


One possible annotation of the input structure is a mask to allow tokens of the question and SPARQL queries to only attend within their segment.
We call this mask as \textit{block attention} and test it as an alternative to the hierarchical attention structures (parse trees).
This mask is denser than the attention mask from parse trees and sparser than ``no mask''.
Figure~\ref{fig:block_attn} shows the block attention for the examples shown in the Figure~\ref{fig:struct_anno}.

Table~\ref{tab:struct_anno_result_full} reports the full results of experiments on structure annotations.
In all cases, entity cross links improve compositional generalization on the dev set, but provide a significant gain only when combined with the parse tree attention and the attention is guided by the ``hard mask''.
As we can see in the ``hard mask'' experiments, block attention does not improve compositional generalization, which suggests a need for more detailed attention mask of input structure.

\begin{table*}[t]
    \setlength{\tabcolsep}{1pt}
    \centering
    \begin{tabular}{c|c|c|c|c|ccc|ccc|ccc} \hline
         \multirow{2}{*}{\textbf{Model}} & \multirow{2}{*}{\shortstack{\textbf{Mask}\\\textbf{Type}}} & \multirow{2}{*}{\shortstack{\textbf{Parse}\\\textbf{Tree}}} & \multirow{2}{*}{\shortstack{\textbf{Block}\\\textbf{Attn}}} & \multirow{2}{*}{\shortstack{\textbf{Cross}\\\textbf{link}}} & \multicolumn{3}{c|}{\textbf{Train}} & \multicolumn{3}{c|}{\textbf{Train (hold-out)}} & \multicolumn{3}{c}{\textbf{Dev}} \\
         & & & & & Acc & F1 & AUC & Acc & F1 & AUC & Acc & F1 & AUC \\ \hline
        LSTM & \multicolumn{4}{c|}{-} & 0.9990 & 0.9979 & 1.0000 & 0.9796 & 0.9604 & 0.9972 & 0.8226 & 0.5199 & 0.8310 \\ \hline
        Transformer & \multicolumn{4}{c|}{-} & 0.9886 & 0.9776 & 0.9995 & 0.9582 & 0.9189 & 0.9931 & 0.8414 & 0.6191 & 0.8738 \\ \hline
        \multirow{9}{*}{\shortstack{Transformer\\ w/ structure\\annotations\\(ETC)}} & No & \multicolumn{3}{c|}{-} & 0.9874 & 0.9751 & 0.9994 & 0.9591 & 0.9199 & 0.9934 & 0.8434 & 0.6202 & 0.8868 \\ \cline{2-14}
        & \multirow{4}{*}{Hard} & Y & N & N & 0.9955 & 0.9911 & 0.9999 & 0.9766 & 0.9543 & 0.9978 & 0.8628 & 0.6744 & 0.9061 \\ \cline{3-5}
        & & Y & N & Y & 0.9978 & 0.9956 & 1.0000 & 0.9866 & 0.9738 & 0.9992 & 0.9170 & 0.8269 & \underline{0.9656} \\ \cline{3-5}
        & & N & Y & N & 0.9828 & 0.9659 & 0.9989 & 0.9567 & 0.9152 & 0.9928 & 0.8324 & 0.5874 & 0.8771 \\ \cline{3-5}
        & & N & Y & Y & 0.9871 & 0.9746 & 0.9993 & 0.9573 & 0.9171 & 0.9930 & 0.8386 & 0.6048 & 0.8881 \\ \cline{2-14}
        & \multirow{2}{*}{Soft} & Y & N & N & 0.9863 & 0.9728 & 0.9993 & 0.9588 & 0.9197 & 0.9933 & 0.8426 & 0.6017 & 0.8729 \\ \cline{3-5}
        & & Y & N & Y & 0.9891 & 0.9784 & 0.9995 & 0.9603 & 0.9226 & 0.9936 & 0.8482 & 0.6385 & 0.8819 \\ \cline{2-14}
        & Hard & Y & N & N & 0.9940 & 0.9882 & 0.9999 & 0.9743 & 0.9500 & 0.9973 & 0.8615 & 0.6697 & 0.9056 \\ \cline{3-5}
        & +Soft & Y & N & Y & 0.9975 & 0.9949 & 1.0000 & 0.9867 & 0.9739 & 0.9991 & 0.9249 & 0.8473 & \textbf{0.9721} \\ \hline
    \end{tabular}
    \caption{Results of the CFQ classification dataset (MCD split \& model negatives) with different types of structure annotations}
    \label{tab:struct_anno_result_full}
\end{table*}

\section{Related Works on Compositional Generalization}
\label{sec:related_works}

In this section, we review prior works on improving compositional generalization in more detail.

\citet{russin2019compositional} proposed to split the attention mechanism into two separate parts, syntax and semantics. The semantic part encodes each token independent of the context (this is a pure embedding look-up table), and the syntactic part encodes each token by looking only at its context (without looking at the token itself). In this way, the syntactic part tries to capture the syntactic role a token might play in a sequence. They show improved compositional generalization on the SCAN dataset using LSTMs, with respect to using standard attention. Compared to \citet{russin2019compositional} that uses LSTMs for the syntactic part, we use Transformer architecture to handle the hierarchical structure of the input.

In their follow up work on the CFQ dataset, \citet{furrer2020compositional} showed that an increased amount of pre-training helped Transformer models better generalize compositionally.

Another idea that has been proposed is to augment the training data, adding synthetic training examples to give the model a compositional learning bias \citep{andreas2019good} .

Finally, work also exists on using general-purpose models like {\em Neural Turing Machines} or {\em Differential Neural Computers}~\cite{graves2016hybrid} that are often trained via reinforcement learning to solve compositional generalization tasks. These models learn an “algorithm” that can solve the task at hand, rather than trying to learn a direct input/output mapping as the Transformer models used in most other works do.

\section{Examples of the CFQ classification dataset}

\begin{figure*}[t]
    \centering
    \begin{subfigure}[b]{\textwidth}
        \centering
        \includegraphics[width=\textwidth]{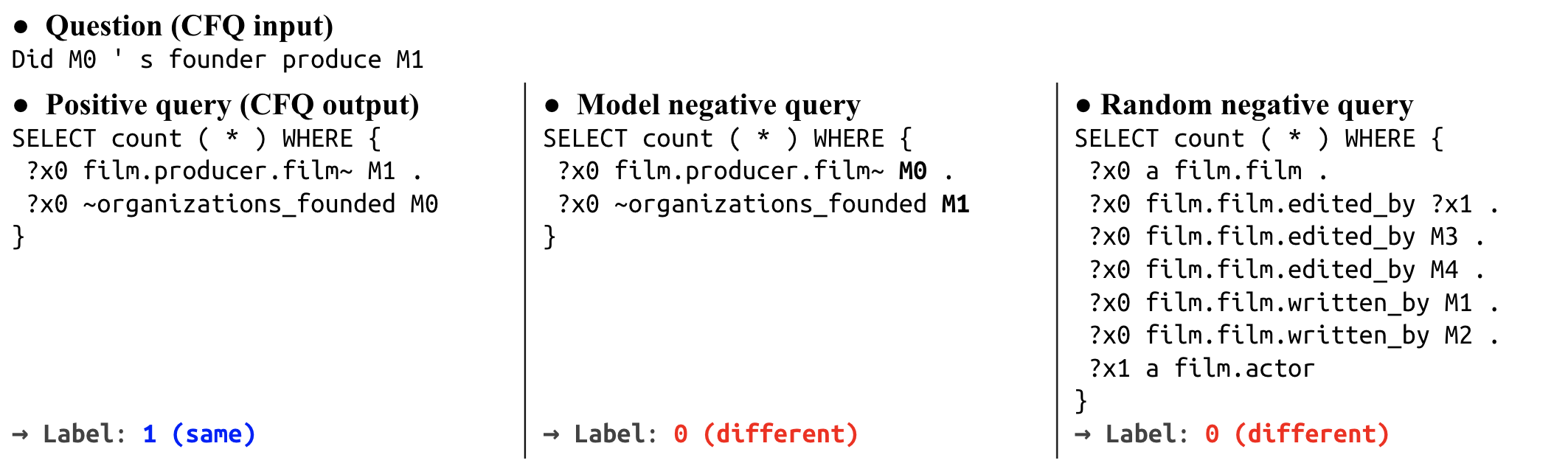}
        \caption{}
        \label{fig:cfq_cls_example_more_1}
    \end{subfigure}
    \begin{subfigure}[b]{\textwidth}
        \centering
        \includegraphics[width=\textwidth]{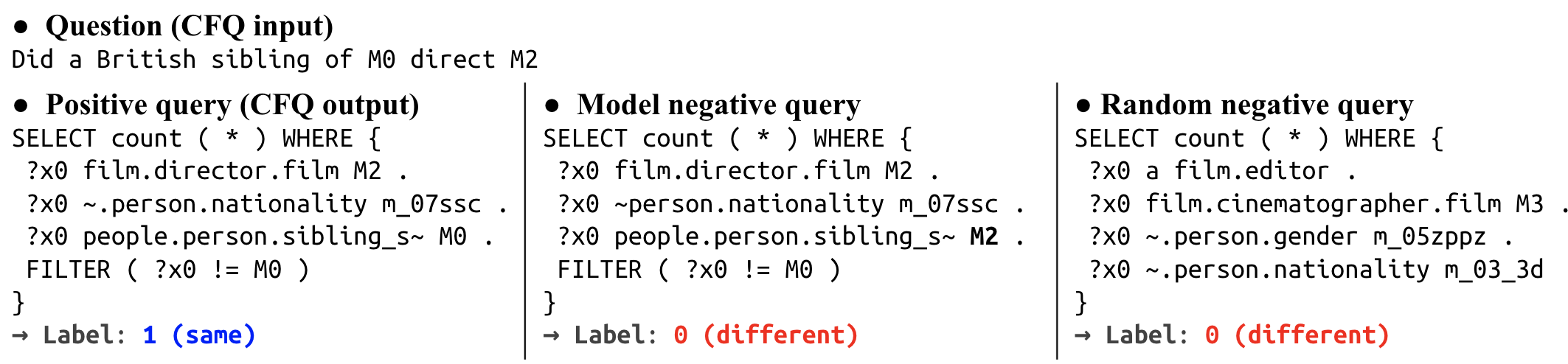}
        \caption{}
        \label{fig:cfq_cls_example_more_2}
    \end{subfigure}
    \begin{subfigure}[b]{\textwidth}
        \centering
        \includegraphics[width=\textwidth]{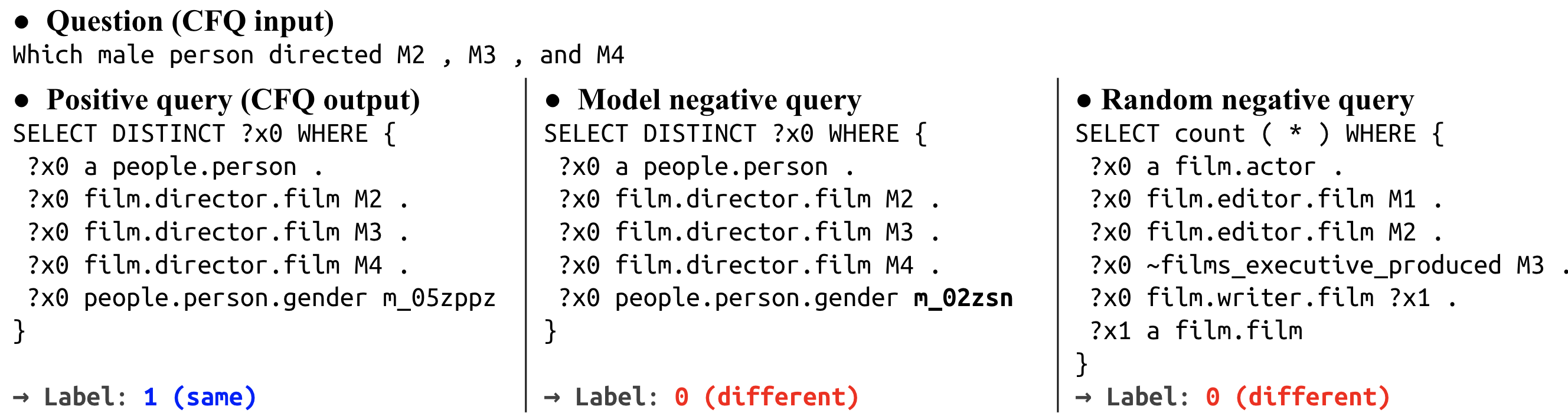}
        \caption{}
        \label{fig:cfq_cls_example_more_3}
    \end{subfigure}
    \begin{subfigure}[b]{\textwidth}
        \centering
        \includegraphics[width=\textwidth]{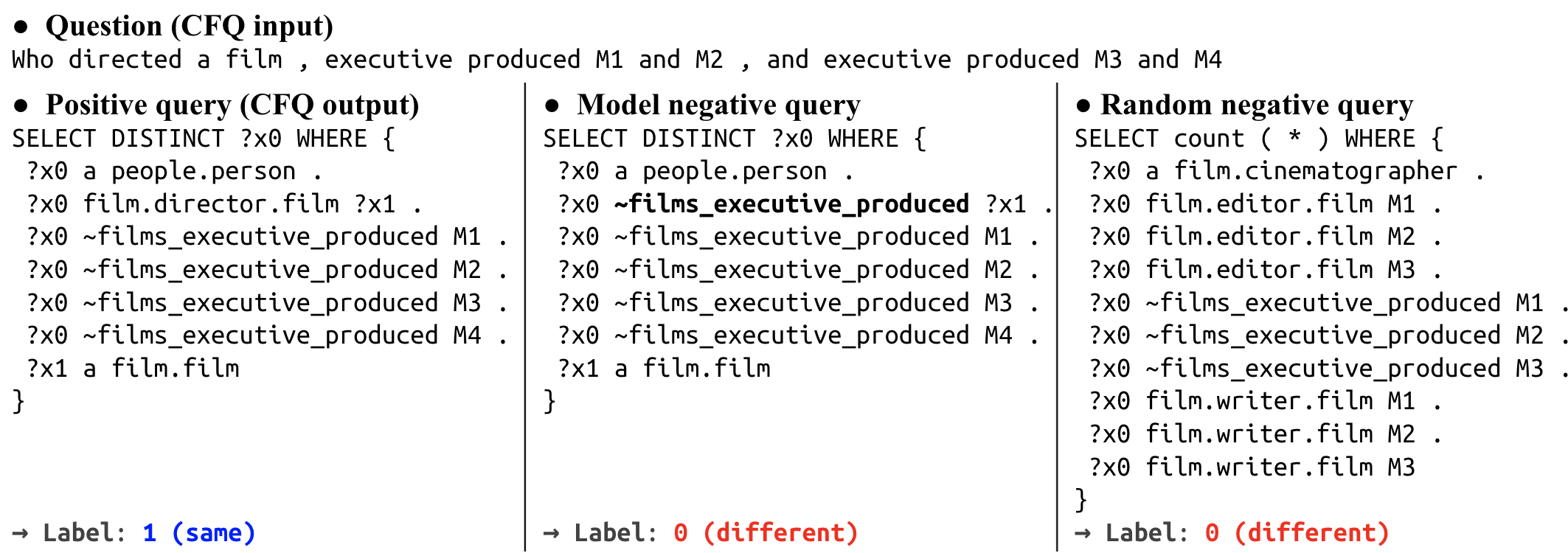}
        \caption{}
        \label{fig:cfq_cls_example_more_4}
    \end{subfigure}
    \caption{Examples of the CFQ classification dataset. Each query pairs with the question to form an instance. Note the model negative resembles the positive, while the random negative query differs considerably. In the model negative queries, the differences from the positive query are marked in bold.}
    \label{fig:cfq_cls_example_more}
\end{figure*}

In Figure~\ref{fig:cfq_cls_example_more}, we present more examples of the CFQ classification datasets.
In all cases, the random negative queries substantially differ from the positive queries, implying that a learner can easily perform the task.
On the other hand, the model negative queries only differ by a token or a phrase, which demands a learner's higher ability.

\end{document}